# AmQA: Amharic Question Answering Dataset

Tilahun Abedissa[♣♥]
tilahun.abedissa@gmail.com

Ricardo Usbeck[♥]
ricardo.usbeck@uni-hamburg.de

Yaregal Assabie[♣]
yaregal.assabie@aau.edu.et

[♣]Addis Ababa University, [♥]University of Hamburg

**Abstract**

Question Answering (QA) returns concise answers or answer lists from natural language text given a context document. To advance robust models' development, large amounts of resources go into curating QA datasets. There is a surge of QA datasets for languages like English, however, this is not the case for Amharic. Amharic, the official language of Ethiopia, is the second most spoken Semitic language in the world. There is no published or publicly available Amharic QA dataset. Hence, to foster the research in Amharic QA, we present the first Amharic QA (AmQA) dataset. We crowdsourced 2628 question-answer pairs over 378 Wikipedia articles. Additionally, we run an XLMR$_{Large}$-based baseline model to spark open-domain QA research interest. The best-performing baseline achieves an F-score of 69.58 and 71.74 in reader-retriever QA and reading comprehension settings respectively.

## 1 Introduction

The task of Question Answering (QA) is to find an accurate answer to a natural language question from a certain underlying data source (Usbeck et al., 2016). To get an as concise answer as possible for a natural language question, a plethora of QA approaches has been proposed (Chen & Yih, 2020). The scientific direction, of curating standard QA datasets is being applied to evaluate models' question synthesis ability, answer accuracy, and stimulate the research in the field (Cambazoglu et al., 2020; Kwiatkowski et al., 2019; Rogers et al., 2021). The existing QA datasets in different languages are commonly curated using either crowdsourcing or automatic generation approaches. In the first approach, crowd-workers formulate question-answer pairs over a given context. This allows for creating high-quality question-answer pairs, but very expensive. In the latter approach, question-answer pairs are formulated using language generation models, machine translation, or manual/learned templates. The main challenge in automatic generation is gold answer extraction. Mostly accomplished using existing QA models. But getting a dependable model, as perfect as a human, that can produce a correct answer is challenging. So, to minimize the generation of trivial and un-grammatical question-answer pairs, aside from improving the performance of the generation models, experts paraphrase the generated question-answer pairs (Cambazoglu et al., 2020).

The distinction between the existing datasets lies in the question types (factoid vs non-factoid) and answer formulation sub-task (extractive vs abstractive). Factoid extractive QA datasets like SQuAD (Rajpurkar et al., 2016), come up with a challenge to measure a QA model competency in identifying the span of an answer from a context for factoid questions. Factoid questions like 'What is the capital city of Ethiopia?' (Answer: Addis Ababa) seeks a factual answer that appears as a named entity such as date, location, proper noun, other short noun phrases, or short sentence. Unlike that, abstractive QA datasets contain questions whose answer is a comprehension of a context, not a direct copy (Fan et al., 2019).

Recently, the QA field of study is getting too many datasets in mono, cross, and multi-lingual settings (Asai et al., 2021; Clark et al., 2020; Gupta et al., 2018; Lewis et al., 2020; J. Liu et al., 2019). However, Amharic[1] is not included yet in the map of the QA datasets. Specific to Amharic there are attempts to develop datasets for other Natural Language Processing (NLP) tasks like sentiment analysis (Yimam et al., 2020), morphologically annotated corpus (Yeshambel et al., 2020), contemporary Amharic corpus (Gezmu et al.,

[1] Amharic is written using Ge'ez script known as ፊደል (Fidel)

**Context**: …በላሊበላ 11 ውቅር ዐብያተ ክርስቲያናት ያሉ ሲሆን ከነዚህም ውስጥ ቤተ ጊዮርጊስ (ባለ መስቀል ቅርፁ) ሲታይ ውሃ ልኩን የጠበቀ ይመስላል። ቤተ መድኃኔዓለም የተባለው ደግሞ ከሁሉም ትልቁ ነው። ላሊበላ (ዳግማዊ ኢየሩሳሌም) የገና በዓል ታህሳስ 29 በልዩ ሁኔታና ድምቀት ይከበራል፣ "ቤዛ ኩሉ" ተብሎ የሚጠራው በነግህ የሚደረገው ዝማሬ በዚሁ በዓል የሚታይ ልዩና ታላቅ ትዕይንት ነው። (While there are 11 rock-hewn churches in Lalibela, of these churches, *betə giorgis* 'House of St. George' (the one that is cross-shaped) appears to have a leveled foundational platform. The church named *betə medhanialəm* (House of the Saviour of the World), is also the biggest of all. In Lalibela (the Second Jerusalem), *gənna* 'Christmas' holiday is celebrated uniquely and colorfully on December 29. The song called *beza kulu* is played in the aftermath of the holiday and it is a great and special scene observed in this holiday.)
**Question**: ከላሊበላ አስራ አንዱ ውቅር አብያተ ክርስቲያናት የመስቀል ቅርጽ ያለው የትኛው ነው? (Of the 11 Lalibela's rock-hewn churches, which one is cross-shaped?)
**Answer**: ቤተ ጊዮርጊስ (*betə giorgis* 'House of St. George')

Figure 1: Sample question from AmQA dataset.

2018), and parallel corpora for machine translation (Abate et al., 2018).

But still, no publicly available dataset can be used for training and/or testing Amharic QA models. In Amharic, interrogative sentences can be formulated using information-seeking pronouns like "ምን" (what), "መቼ" (when), "ማን" (who), "የት" (where), "የትኛው" (which), etc. and prepositional interrogative phrases like "ለምን" [ለ-ምን] (why), "በምን" [በ-ምን] (by what), etc. Besides, a verb phrase could be used to pose questions (Getahun 2013; Baye 2009). As shown in Figure 1, the AmQA dataset contains context, question, and answer triplets (also see Figure 3 in Appendix A). The contexts are articles collected from Amharic Wikipedia[2]. The question-answer pairs are created by crowd workers using the Haystack [3] QA annotation tool. 2628 question and answer pairs are created from 378 documents. For example, for the question given in Figure 1, the answer is the span ቤተ ጊዮርጊስ (betə Giorgis 'House of St. George') from the context. In our work, in addition to the crowd-sourced question-answer pairs, we have set baseline F1-score values by implementing a QA model with the retriever and reader components.

Generally, given the QA dataset, it can be used to test end-to-end QA and reading comprehension models. Besides that, it is suitable for testing retriever-reader pipelined QA models using the contexts as they are or in a full-Wikipedia setting. That makes the AmQA dataset used as an open-domain QA resource. Thus, our contribution is the first Amharic, open-domain QA dataset which will hopefully foster the development of better QA approaches for Amharic. The dataset can be found online at https://github.com/semantic-systems/amharic-qa.

## 2 Related Works

Among the existing English QA datasets, SQuAD (Rajpurkar et al., 2016, 2018) paved the way by creating question-answer pairs from Wikipedia articles using crowd workers, where each question answer is a span of text in the articles. Following the SQuAD footsteps, Chinese MRC (Cui et al., 2019), Vietnamese QA (Do et al., 2021), and other datasets listed in (Dzendzik et al., 2021; Rogers et al., 2021) are created with a little distinction on curation steps. On the other hand, by automatically translating SQuAD into their respective languages German (Möller et al., 2021), French (Hoffschmidt et al., 2020), and Arabic (Mozannar et al., 2019) versions are created. Translating existing QA datasets to other languages would be a mighty solution to have a dataset with a large size. However, due to the lack of well-tested open-access automatic translators, we couldn't use this approach.

For Amharic, there are very few QA models, TETEYEQ (Yimam & Libsie, 2009) answers factoid-type questions by extracting entity names using a rule-based answer extraction approach. Abedissa & Libsie (2019) introduced a non-factoid QA model that answers biography, description, and definition questions. The definition-description answer extraction is accomplished using heuristics. Whereas biography questions are answered using a summarizer and a classifier to determine whether the summary is a valid biography or not. Both works, beyond the attempt to answer Amharic questions, didn't produce a published dataset that can be used to test the performance of Amharic QAs.

[2] https://am.wikipedia.org/wiki/ዋናው_ገጽ

[3] https://docs.haystack.deepset.ai/docs/annotation

The lack of standard public Amharic QA datasets along with the scarcity of different add-in Amharic Natural Language Processing (NLP) tools like part-of-speech-tagger, stemmer, anaphora resolver, etc. hindered the development of Amharic QA approaches. Hence, in this work, we provide an AmQA data set that can be used as a testbed for Amharic QA models as well as cross-lingual and/or multi-lingual QA models.

## 3 The AmQA Dataset

The AmQA dataset is created following three phases: article gathering, crowdsourcing question-answer pairs, and question-answer pair validation.

### 3.1 Article collection and cleaning

The Amharic articles used as contexts are collected from the Amharic Wikipedia dump[4] file and those articles whose sizes are greater than 2 KB are kept. Articles under the 'proverb' and 'food preparation' categories are removed. Proverb articles are favorable for creating reasoning questions. Besides, 'food preparation' articles mostly contain steps of the preparation of food, which are suitable for creating 'how is the step …', and 'list the steps | ingredients added to …' questions. In both cases, even the answer may not be a span of a text in the article. The remaining articles after filtration are further pre-processed by the wiki_dump_reader[5] tool to get clean texts. At last, since long articles do not motivate to create questions exhaustively, each article is chunked using the sub-topics in it. Then, we randomly select 378 cleaned articles.

### 3.2 Question-Answer Pair Crowdsourcing

In the question-answer pair formulation, the cleaned contexts along with sample examples are distributed to native Amharic speaker crowd workers who have at least Bachelor's degree. Training[6] is given on how to create questions that can be answered in each context. Since the articles are randomly selected from Wikipedia, the crowd workers are advised to report when they found an article with offensive content. The crowd workers are free to formulate as many questions as possible from a given context.

|  | Article | Question | Answer |
|---|---|---|---|
| size | 378 | 2628 | 2628 |
| word len (avg) | 172.07 | 9.22 | 2.66 |

Table 1: Sample question from AmQA dataset.

### 3.3 Question-Answer Pair Validation and Annotation

The validation of the formulated question-answer pairs is about their correctness and completeness. When we say correctness, the posed questions should be answerable by the given context and their answer should be precise. For example, a question like 'How many parks are there in our country?', is ambiguous due to the possessive adjective 'our', such questions are paraphrased according to the context. Questions that do not explicitly state the subject/object are paraphrased. Ambiguous, too long, and questions with non-consecutive string answers are excluded from the annotation. Then, the validated question-answer pairs are annotated using the Haystack [7] annotation tool. The annotation tool provides the annotated question-answer pairs as JSON files in SQuAD format. Since the annotator introduces the '\n' character in the exported file, it is removed.

## 4 Dataset Analysis

This section presents the analysis of the dataset. Also, provides different statistics that show the features of the dataset.

### 4.1 Data statistics

Table 1 shows the number of articles, questions, and answers along with the average word length of documents, questions, and answers. The contexts in the AmQA dataset on average contain 172 words. Most questions' average word length is 9.22; whereas the answers are short, and their average word length is 2.66.

### 4.2 Questions Expected Answer Type

To compute the percentage of the expected answer types 300 questions are selected randomly. Then, the questions are categorized into a person, location, time, organization, number, description, and other classes based on the interrogative terms and the answer phrase. As shown in Table 2, we found that most of the questions are about Location, Number, and Time, where each type has above 18% coverage. Description questions take 13% of the share and questions that look for a person's name as an answer are 14.38%. 10.7% of questions, expected answer type are entities that cannot be included in the existing categories, and fall into the 'OTHER' group. Among the questions, list (3.01%) and organization (2.67%) are the smallest. In addition, Figure 2 (See Appendix A), shows the distribution of the interrogative terms over the randomly selected questions.

[4] https://dumps.wikimedia.org/amwiki/20210801/ last accessed Aug. 18, 2021

[5] https://pypi.org/project/wiki-dump-reader/

[6] We follow the guideline given in the annotation tool handbook.

[7] Haystack Annotation Tool (deepset.ai)

| EAT | % | Example |
|---|---|---|
| Person | 14.38 | አዲስ አበባ የሚለውን ስም ለከተማዋ የሰጡት ማናቸው? (Who gave the name Addis Ababa to the city?) |
| Location | 18.72 | የአፍሪካ ሕብረት መቀመጫ መዲና ማናት? (Which city is the place of Africa Union? ) |
| Time | 18.06 | ሉሲ በኢትዮጵያ የተገኘችው መቼ ነበር? (When was Lucy found in Ethiopia?) |
| Organization | 2.67 | ቀዳማዊ ኃይለ ሥላሴ ዩኒቨርሲቲ አሁን ምን በመባል ይጠራል? (What is the current name of Haile Silasie I University?) |
| Number | 18.72 | በጣና ሐይቅ ምን ያህል ደሴቶች አሉ? (How many islands are there in lake Tana?) |
| Description | 13.71 | ውክፔዲያ ምንድን ነው? (What is Wikipedia?) |
| List | 3.01 | የጋና ዋና ምርቶች ምንድን ናቸው? (What are the main products of Ghana?) |
| Other | 10.7 | የኢትዮጵያ የስራ ቋንቋ ምንድን ነው? (What is the working language of Ethiopia?) |

Table 2: Sample question from AmQA dataset.

## 5 Experiment

### 5.1 Baseline Model

Since the AmQA dataset contains a set of contexts along with question-answer pairs, it can be considered a reading comprehension (RC) task (Dzendzik et al., 2021; Lewis et al., 2020). That is, given a question Q and a context consisting of words, the goal of the model is to identify a word or group of consecutive words that answers question Q. Hence, based on this assumption we have set a baseline value for the AmQA using XLM-R (Conneau et al., 2020) based QA model that was fine-tuned on SQuAD 2.0 dataset (Rajpurkar et al., 2018). The Cross-Lingual Language Model-RoBERTa (XLM-R) is a multilingual pre-trained transformer model based on the RoBERTa architecture and trained using 2.5 TB of data across 100 languages including Amharic (Conneau et al., 2020; Y. Liu et al., 2019).

On the other hand, since retriever-reader-based QA models first retrieve relevant passages, then read top-ranked passages and try to predict the start and end positions of the answer, we have implemented a retriever-reader (RR) QA model using the Farm Haystack[8] open-source framework. For the retriever part, we have used BM25 and XLM-R$_{Large}$[9] is used as a reader.

| Settings | EM | F1 |
|---|---|---|
| XLM-R performance on MLQA (Conneau et al., 2020) | 52.7 | 70.7 |
| RC(XLM-R$_{base}$[1]) | 47.49 | 64.69 |
| RC (XLM-R$_{large}$) | 50.76 | **71.74** |
| RR QA | 42.4 | 64.3 |
| RR QA + Pre-Processor | 50.3 | **69.58** |

Table 3: XLM-R based models' performance for RC and RR settings oven the AmQA dataset.

### 5.2 Evaluation

The evaluation stage measures the performance of a QA model (F-Score), the accuracy of the returned answers (EM), as well as the difficulty level of a QA dataset (Clark et al., 2020; Kwiatkowski et al., 2019; Usbeck et al., 2016). A baseline value over the AmQA dataset is computed using F-score and exact match (EM) metrics.

As shown in Table 3, on the reading comprehension setting the XLM-R$_{Large}$ F1 score is 71.74, whereas the XLM-R$_{Base}$ F1 score is 64.69. This shows that XLM-R$_{Large}$ performs better than the XLM-R$_{Base}$ model. In addition, since the F1 score of the XLM-R$_{Large}$ on the AmQA dataset is comparable to the average F1 score of the XLM-R on the MLQA dataset for seven different languages (70.7), we have decided to use it as a reader component in the RR QA model. From our observation, we have noticed that some returned answers by the models contain the gold answer but have affixes, additional strings, unnecessary blank spaces, and/or punctuations. So, we have created a pre-processor that normalizes characters and removes punctuation, quotation marks, and spaces. As a result, the RR QA model shows some improvement with the pre-processor.

[8] https://haystack.deepset.ai/

[9] https://huggingface.co/deepset/xlm-roberta-large-squad2

## 6 Summary & Outlook

In this paper, we presented an Amharic Question Answering dataset that contains triplets of documents, questions, and answers curated using Amharic Wikipedia. In addition, we have set baseline values in reading comprehension and retriever-reader settings. We hope the introduction of the AmQA dataset will stimulate researchers to test monolingual and/or multilingual QA models. Besides, if the equivalent translation of the curated data is obtained, this data can be used for cross-lingual QA models.

## Limitations

AmQA is only a small dataset due to the expensive labor involved in creating it. Thus, data-intensive methods are disadvantaged. Also, the annotations were done by a limited number of human annotators and thus may have inherent biases or systematic annotation errors. We will investigate this in future funded work on low-resource languages. Also, the choice of baselines was limited by available computing resources. There might be better out-of-the-box baselines, such as Huggingface's Bloom, which perform better.

Appendix A

Figure 2: Interrogative terms distribution in AmQA dataset

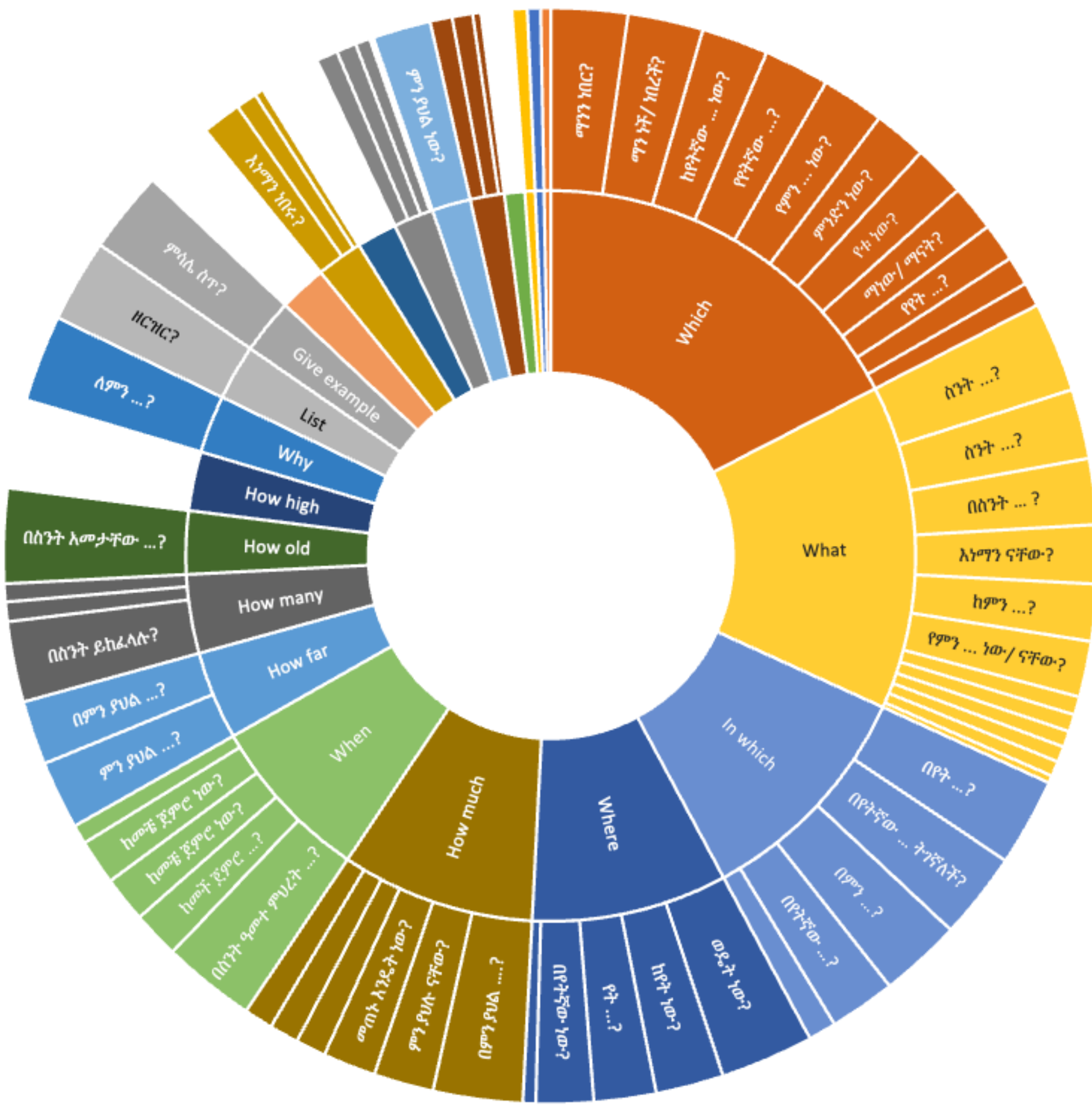

Figure 3: AmQA dataset SQuAD like example

```
            {
              "question": "በላሊበላ የጌታ ልደት ቀን በቤተ ማርያም የሚቀርበው ልዩ ዝማሬ ምን ይባላል?",
              "id": 272836,
              "answers": [
                {
                  "answer_id": 270480,
                  "document_id": 266719,
                  "question_id": 272836,
                  "text": "ቤዛ ኩሉ",
                  "answer_start": 465,
                  "answer_end": 470,
                  "answer_category": null
                }
              ],
              "is_impossible": false
            }
          ],
          "title": "ላሊበላ",
          "context": "ንጉሡ ላሊበላ የሚለውን ስም ያገኘው፣ ሲወለድ በንቦች ስለተከበበ ነው ይባላል። ላል ማለት ማር ማለት ሲሆን፤ ላሊበላ
ማለትም -ላል ይበላል (ማር ይበላል) ማለት እንደሆነ ይነገራል።  ውቅር ቤተክርስቲያናቱን ንጉሡ ጠርቦ የሰራቸው ከመላእክት እገዛ ጋር እንደሆነ በኢትዮጵያ
ኦርቶዶክስ እምነት ተከታዮች ይነገራል። በ16ኛው ክፍለ ዘመን አውሮፓዊ ተጓዥ ላሊበላን ተመልክቶ «ያየሁትን ብናገር ማንም እንዴኔ ካላየ በፍጹም
አያምነኝም» ሲል ተናግሮ ነበር። በላሊበላ 11 ውቅር ዐብያተ ክርስቲያናት ያሉ ሲሆን ከነዚህም ውስጥ ቤተ ጊዮርጊስ (ባለ መስቀል ቅርፅ) ሲታይ ውሃልኩን
የጠበቀ ይመስላል። ቤተ መድሃኔ ዓለም የተባለው ደግሞ ከሁሉም ትልቁ ነው። ላሊበላ (ዳግማዊ ኢየሩሳሌም) የገና በዓል ታህሳስ 29 በልዩ ሁኔታ ና
ድምቀት ይከበራል፣ \"ቤዛ ኩሉ\" ተብሎ የሚጠራው በነግህ የሚደረገው ዝማሬ በዚሁ በዓል የሚታይ ልዩ ና ታላቅ ትዕይንት ነው።የሚደረገውም ከቅዳሴ
በኋላ በቤተ ማርያም ሲሆን ከታች ባለ ነጭ ካባ ካህናት ከላይ ደግሞ ባለጥቁር ካብ ካህናት በቅዱስ ያሬድ ዜማ ቤዛ ኩሉ እያሉ ይዘምራሉ። 11ዱ የቅዱስ
ላሊበላ ፍልፍል አብያተ ክርስቲያናት ቤተ መድሃኔ ዓለም፣ ቤተ ማርያም፣ ቤተ ደናግል፣ ቤተ መስቀል፣ ቤተ ደብረሲና፣ ቤተ ጎለጎታ፣ ቤተ አማኑኤል፣ ቤተ አባ
ሊባኖስ፣ ቤተ መርቆሬዎስ፣ ቤተ ገብርኤል ወሩፋኤል፣ ቤተ ጊዮርጊስ ናቸው።",
          "document_id": 266719
        }
```